\documentclass{article}

\usepackage[final]{nips_2017}

\usepackage[utf8]{inputenc} 
\usepackage[T1]{fontenc}    
\usepackage{hyperref}       
\usepackage{url}            
\usepackage{booktabs}       
\usepackage{amsfonts}       
\usepackage{nicefrac}       
\usepackage{algorithm}
\usepackage[noend]{algpseudocode}
\usepackage{float}
\usepackage{amsmath}
\usepackage{amssymb}
\newcommand{\R}{\mathbb{R}}
\usepackage{graphicx,pgfplots,verbatim,adjustbox,amsmath}            \usepackage{mathtools}
\usepackage{algorithm}
\usepackage[noend]{algpseudocode}

\algrenewcommand\algorithmicforall{\textbf{foreach}}
\algrenewcommand\algorithmicindent{.8em}
\usepackage{array}
\usepackage{eqparbox}

\usepackage{array}
\usepackage{eqparbox}

\usepackage{microtype}      

\makeatletter
\def\BState{\State\hskip-\ALG@thistlm}
\makeatother
\makeatletter
\usepackage[noend]{algpseudocode}
\usepackage{float}
\algnewcommand\INPUT{\item[\textbf{Input:}]}%
\algnewcommand\OUTPUT{\item[\textbf{Output:}]}

\author{
  Housam Khalifa Bashier Babiker and Randy Goebel\\
Alberta Machine Intelligence Institute\\
  Department of Computing Science University of Alberta\\
  Edmonton, Alberta Canada T6G 2E8 \\
  \texttt{khalifab@ualberta.ca,rgoebel@ualberta.ca} \\
}

\title{An Introduction to Deep Visual Explanation}

\title{An Introduction to Deep Visual Explanation}

\begin{document}

\maketitle

\begin{abstract}
The practical impact of deep learning on complex supervised learning problems has been significant, so much so that almost every Artificial Intelligence problem, or at least a portion thereof, has been somehow recast as a deep learning problem.  The applications appeal is significant, but this appeal is increasingly challenged by what some call the challenge of explainability, or more generally the more traditional challenge of debuggability: if the outcomes of a deep learning process produce unexpected results (e.g., less than expected performance of a classifier), then there is little available in the way of theories or tools to help investigate the potential causes of such unexpected behavior, especially when this behavior could impact people's lives.

We describe a preliminary framework to help address this issue, which we call "deep visual explanation" (DVE).  "Deep," because it is the development and performance of deep neural network models that we want to understand.  "Visual," because we believe that the most rapid insight into a complex multi-dimensional model is provided by appropriate visualization techniques, and "Explanation," because in the spectrum from instrumentation by inserting print statements to the abductive inference of explanatory hypotheses, we believe that the key to understanding deep learning relies on the identification and exposure of hypotheses about the performance behavior of a learned deep model.

In the exposition of our preliminary framework, we use relatively straightforward image classification examples and a variety of choices on initial configuration of a deep model building scenario.  By careful but not complicated instrumentation, we expose classification outcomes of deep models using visualization, and also show initial results for one potential application of interpretability. 
 
\end{abstract}

\section{Introduction}

The primary appeal of deep learning is that a predictive model can be automatically constructed from a suitable volume of labeled inputs.  In an increasing number of demonstration applications, the staging of a deep learning exercise need only outline the details of the supervised learning problem in terms of input data, and leave the creation of the predictive classifier to the deep learning framework (e.g., Google's Tensorflow, Microsoft CNTK).  The fundamental improvement of current deep learning methods is that, unlike earlier more shallow network layers, deep learning automatically identifies appropriate stratification of a predictive model [5]. This property of finding appropriate multi-layer structures of a supervised classification problem has produced significant advances in AI systems, especially those that rely on accurate classification, including automated driving, voice recognition, some natural language processing tasks, and image classification.

Because many components of Artificial Intelligence systems include classification components, it is easy to imagine that the construction of accurate classification components provide an essential contribution to overall intelligent systems.  When classifiers are simple and the categories are well-defined (e.g., classifying humans by sex), then it is relatively easy to confirm whether a classifier is performing well.  But when the classification is more complex, e.g., classifying complex proteins and their potential docking targets into potentially active pairings, then it is not so easy to determine what a deep learned classifier is doing, especially if unexpected pairs are predicted. 

It is not a surprise that, for as long as programming has been done (by humans or machines), there has always been the need for supporting systems that help programmers understand the unexpected behaviour from their programs.  From primitive but effective old ideas like the insertion of print statements, to the sophistication of non-monotonic abductive hypothesis management systems, the motivation has always been to instrument the computational object of interest to reveal local behaviour and provide insight into whether the "unexpected" outputs are unanticipated insights, just bugs, or some unintended modelling bias. 

What we do here is make some obvious heuristic choices about how to instrument deep learned models, and then assemble a collection of components to provide a suggestion about how to approach the idea of deep visual explanation (DVE).  The DVE name arises because (1) our focus is on understanding the scope of methods that would potentially provide insight into the "black box" of deep-learned models, (2) "visual," because we already believe that the trajectory of successful applications of deep learning are sufficiently complex so that simply identifying some human-manageable small set of parameters will not provide sufficient insight (thus we need visualization methods to help provide insights on multi-dimensional spaces), and (3) "explanation," because we expect that deep-learned models are necessarily always partial, and that there will always be competing alternative explanatory hypotheses about unexpected behaviour.

In the following, we explain our idea by describing a general method to instrument a deep learning architecture,  in this case  an example of a deep learned model representation called VGG-16 networks  [13].  Note that, if we can demonstrate the value of DVE on one such deep representation, we would expect to generalize the framework for a variety of other such deep neural network learning representations.

Our method proposes the creation of a series of multi-dimensional planes that "slice" a multi-layered deep-learned model, so that a few examples of methods of which learned-model attributes could be selected and displayed in a visualization plane, to provide insight into the overall classification performance of the deep-learned model.  Our description does not exhaust the alternatives for how to select visualization techniques or identifying multi-level attributes that provide the "best" insight.  Rather, like in all principled visualization methods,  we don't think there are single best methods for doing that.   Instead, we want to describe a general idea that can be refined in a variety of ways, from a variety of existing literature (including visualization and abductive hypothetical reasoning), in order to create the framework to support the understanding of deep-learned models and their alternatives. 

The reminder of this paper is organized as follows. Section 2 reviews some related work.  Section 3 presents our proposed approach. In Section 4 we describe  the experiments with our initial framework,   and finally, Section 5 concludes with our preliminary results, followed by a discussion of extensive future work.

\section{Ideas from Some Related Work}
Over the last few years, deep learning algorithms have shown impressive results on variety of classification problems such as image classification [16,9], video action recognition[2], natural language processing [11,15], object tracking [8], image segmentations [12] and many others. But designing such a network architecture is a challenging task, especially in the case of trying to understand performance. Many questions are encountered: e.g.,  when things don't work, why is performance is bad? What decision led to a classification outlier?  Why was one class predicted over another? How can one debug an observed error? Why should an output be trusted?

A few  methods have been recently proposed to address these kinds of questions. One approach is to analyze how a deep network responds to a specific input image for a given prediction  [3].  A recent probabilistic method is proposed in [19], the authors assign a weight to each feature with respect to class $y$. The drawback of this approach is that, it is computationally expensive.  Other algorithms are proposed in  [7,14,17, 20].

Another interesting type of explanation is based on network activation; two popular methods are proposed here.
The first method is  called "class activation mapping" (CAM) [18]. The main idea is to estimate the discriminative localization map employed by a CNN configuration. CAM computes a linear relation between feature maps and the weights of the final layer. However, a drawback of this approach is that it can not be employed to networks which use multiple fully connected layers. Recently, in [10], authors proposed a relaxation generalization to CAM known as (GRAD-CAM). It computes the gradients of the predicted class  with respect to feature maps  to obtain weights. The weights are then multiplied with the last pooling layer to identify a discriminative region.

\section{Deep visual explanation (DVE)}
Deep convolutional neural networks (DCNN) produce spatial information at the convolution layers. However, this  information is lost when propagating it to the fully connected layers. This loss of  information makes the explanation process  challenging, especially when it comes to interpreting the output of sensitive data such as medical images.

\begin{figure}[H]
\centering
\includegraphics[width=10cm,height=4cm]
{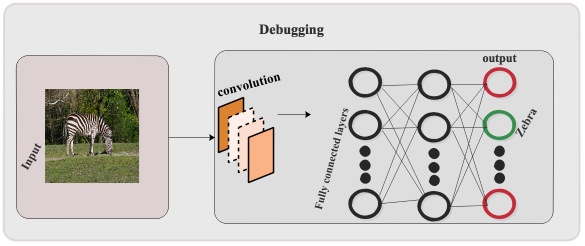}
\caption{Overview of the problem: our goal  is to be able to answer the question of why you arrived at a particular decision?.}
\label{fig:debugging}

\end{figure}

While we recognize that explanations will have many different representations (e.g., image components, language segments, speech segments, etc.), our demonstration here is intended to be simple and preliminary, to illustrate the idea.
Our immediate goal is to create an explanation about the outcome of a DCNN, i.e., to identify which discriminative pixels in the image influence the final prediction (see Figure \ref{fig:debugging}.)\par

\begin{figure}[H]

\centering
\includegraphics[width=10cm,height=4cm]
{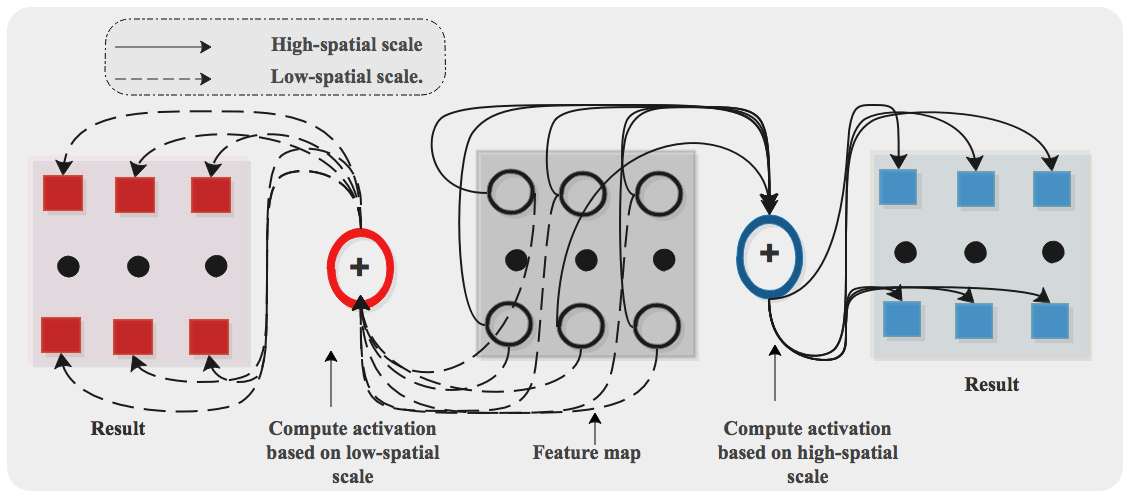}
\caption{Computing the activation of low and high spatial scale for every feature map. The red-circle and blue-circle represent the computation function. }
\label{fig:activation}
\end{figure}

To approach this task in this restricted context,  we assume that the convolution feature maps $X$ at pooling layer $l$ contain some relevant information about class $y_{i}$. We can then write our solution as: $D : I \rightarrow y_{i} \rightarrow S$ i.e., map the input $I$ to class $y_{i}$ using network $D$, and compute the evidence/explanation $S$. Generally, an explanation should be composed of some fragments of features that are crucial in producing the classification output.  So to explain $y_{i} \rightarrow S$,  we can compute the low-spatial scale and high-spatial scale  activations of every feature map, as shown in Figure \ref{fig:activation}. We use the term "activation" here, because we are looking for  those pixels which activate to create either the high or low spatial scale computations. 

Because our explanation here  is in the Fourier domain, we consider a function $F(x)$ representing the transformation where $x \in \R$ and $x$ is a feature map at a particular convolution layer. Therefore, the transformation of every $x_{i} \in X $ of size $MxN$ into Fourier domain  can be written as follows:
\begin{equation}
\begin{split}
 F(u,v) & = \sum_{k=0}^{M-1} \sum_{j=0}^{N-1} f(k,j)e^{-i2\pi(\frac{uk}{M}+ \frac{vj}{N})} \\
 \end{split}
\end{equation}
Where $f(k,j)$ represents a feature map  at layer $l$, the exponential term represents the basis function and the inverse of Fourier is defined as follows:
\begin{equation}
\begin{split}
   f(m,n) & = \frac{1}{MxN} \sum_{u=0}^{M-1} \sum_{v=0}^{N-1} F(u,v)e^{i2\pi(\frac{ux}{M}+ \frac{vy}{N})} \\
 \end{split}
\end{equation}

For every feature map $x_{i} \in X$, we can therefore estimate the visual explanation as: 

\begin{equation}
\begin{split}
 S & = \sum_{i=1}F^{-1}(F(x_{i})*G_{1})* F^{-1}(F(x_{i}*(1-G_{2})) \\
 \end{split}
\end{equation}
Where $G_{1}, G_{2}$ are Gaussians computed at different $\sigma$, $F$ represents the transformation  into Fourier space and $F^{-1}$ denotes the inverse.Equation $3$, computes two types of activations i.e., low-spatial scale activation $(F(x_{i})*G_{1})$ and high-spatial scale activation $(F(x_{i}*(1-G_{2}))$ in Fourier space. 
The advantage of this approach is that, the spatial frequency coefficients are not abruptly cut, but exhibit a gradual cut;  this is essential in order to preserve the discriminative pixels. \par

\begin{figure}[H]

\centering
\includegraphics[width=14cm,height=6.5cm]
{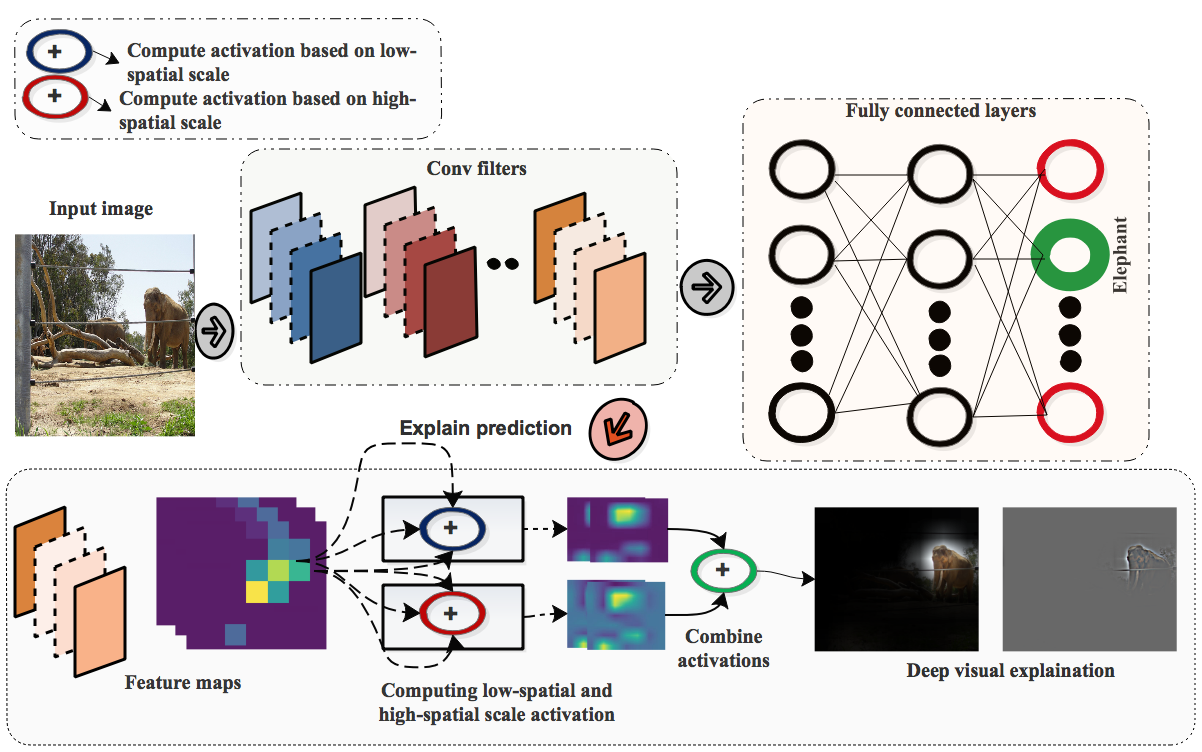}
\caption{Our proposed framework: the input image is passed to the network for class prediction. Once a classification decision is obtained, the explanation is computed from the last pooling layer.}
\label{fig:main_d}
\end{figure}

After computing the visual explanation $S$, we observed that some activations do not contribute in explaining the decision and we refer to this problem as noisy activations. To address it,  we use  $(4)$ to filter out the noise:
\begin{equation}
\begin{split}
 S =  S / \Bigg[ \frac{1}{1+((-1*(S \cdot S^{T})+ V)^{T}+V)
} \Bigg ]
 \end{split}
\end{equation}
Where $V$ is defined as $(1,1,1,...,1)S^{2}$. By using $(4)$, we can highlight the features which contributed substantially to the classification. The overall methodology is depicted in Figure \ref{fig:main_d} and Algorithm \ref{algorithm1} summarizes the overall process.

\begin{algorithm}
\caption{Deep visual explanation}\label{algorithm1}
\begin{algorithmic}
\INPUT image $I$
\OUTPUT Discriminative localization map $S$
\State $Y^{i} \gets \textit{Estimate the highest class score for $I$}$ 
\State $X \gets \textit{Select feature maps}$
\State $S \gets \textit{Intialize to zeros of size $MxN$}$

\For{$j = 1$ \textbf{to} ${nFeatureMaps}$}
\State $S_{temp}$ $\gets \textit{$ Explain(x_{j})$  using eq \color{red} $3$} $
	\State S $\gets \textit{$S + (S_{temp}/K)$  where $K$  is obtained using eq \color{red} $4$ } $
\EndFor

\BState {\textbf{end for}}

\end{algorithmic}
\end{algorithm}

\subsection{Targeted deep visual explanation}
In our simple case of image classification (cf. speech, language) one of the ultimate goals of the visual explanation in the context of debugging is to be precise when determining the component salient patch. Therefore, we should penalize any  activations that do not contribute much in Algorithm \ref{algorithm1}. To handle this,  we propose a method called targeted-DVE to provide a more  targeted explanation. This algorithm removes  any pixel that has less influence on the best explanation.  The process is identical to our previous approach except that, we slightly modify the final output $S$ obtained in Algorithm \ref{algorithm1}. This is done, by  computing $S^{'}$ as follows: 
\begin{equation}
\begin{split}
 S^{'} & = F^{-1}(F(S)*G_{1})* F^{-1}(F(S*(1-G_{2})) \\
 \end{split}
\end{equation}

\section{Experiments}
Here we evaluate our visualization in the context of DCNN classifications. We used images from common objects in context (COCO) challenge set [6], which consists of $91$ objects types. For our example network model, we adopted the publicly available pre-trained VGG-16 network [13].   \par

\textit{Explaining DCCN predictions}

\begin{figure}[H]
\centering
\includegraphics[width=14cm,height=6cm]
{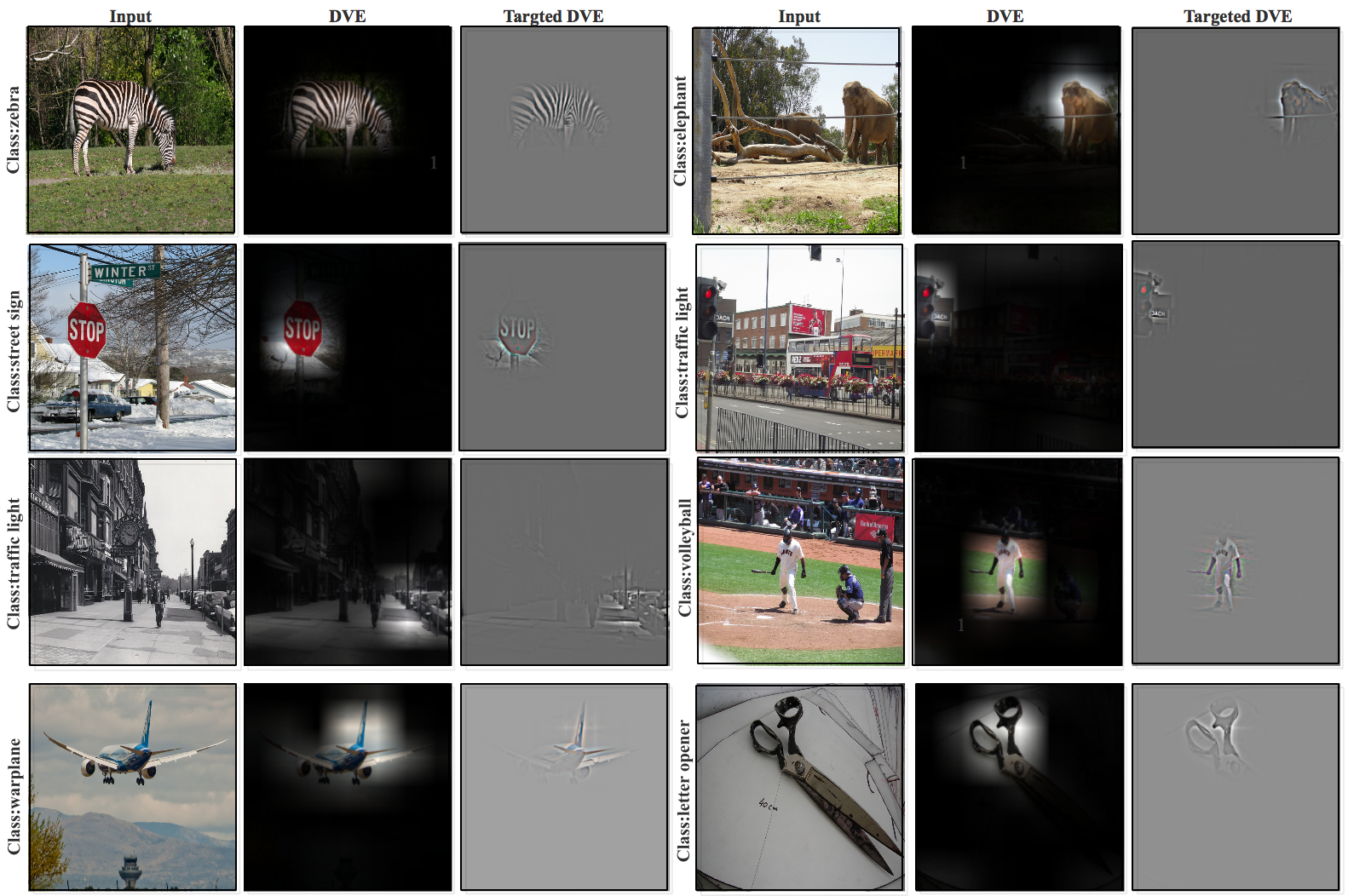}
\caption{Explaining the decisions made by the VGG-16(1 refers to the DVE and 2 refers to targeted-DVE). The network makes correct predictions (zebra, elephant, street sign and traffic light), and our algorithm  provides improved targeted explanation, i.e. it highlights the most discriminative pixels employed by the network. We also show explanations for the incorrect predictions (traffic light, volleyball, warplane, and letter opener). }
\label{fig:prediction}

\end{figure}

\begin{figure}[H]
\centering
\includegraphics[width=14cm,height=7cm]
{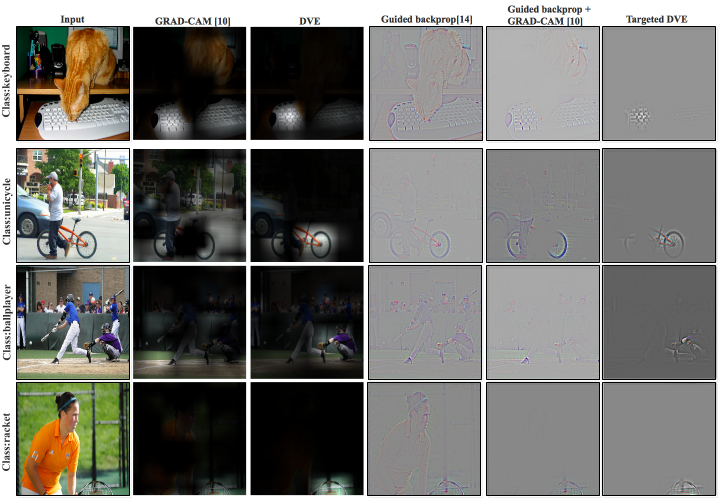}
\caption{Visualizing the VGG-16 network predictions (keyboard, unicycle, ballplayer and racket), and the comparison with other methods.We can see that the generated visualizations of our approach are clearly interpretable.  }
\label{fig:comp}

\end{figure}

The results on randomly selected images from COCO using VGG-16 classifications are depicted in Figure \ref{fig:prediction} and the comparison with other methods is illustrated in Figure \ref{fig:comp}. 

Our approach, does not require training or changing the network architecture. The model also does not require  solving any optimization problem as in [19]. Moreover, our approach is computationally efficient and the computation time on Intel Core $i7$ CPU at $3.60$ GHz is $6^{-4}$ seconds. Finally the algorithm, identifies a relatively minimal discriminative/salient patch that impacts the output of the network.\par
\begin{figure}[H]
\centering
\includegraphics[width=13.5cm,height=4cm]
{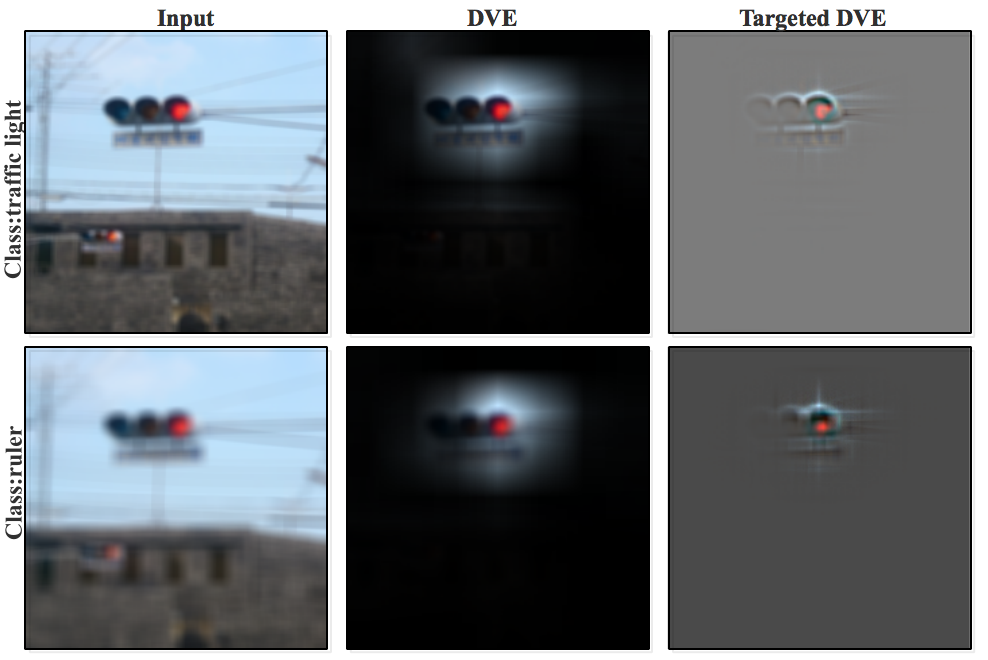}
\caption{This Figure shows the responds of the network to Gaussian blur. The image on the top has less blurring affect than the one on the bottom. We can see that the network is still able to predict the right class (traffic  light), however the network failed to correctly predict the class of the image on the bottom. }
\label{fig:blur}
\end{figure}

\textbf{How does the network see images?}: We also evaluated the robustness of the algorithm against blurring affect. We blurred the image using a Gaussian blur at different $\sigma$, see Figure \ref{fig:blur}. The result suggests that the network is not able predict the blurred images correctly (as we increase $\sigma$) even though the network is looking at the right region. This  means that the network is only looking for specific features in the image (strictly relying on certain pixel values) and is therefore not resistant against blurring effects.  

\textit{Understanding how DCNN compresses information}

\begin{figure}[H]
\centering
\includegraphics[width=13.5cm,height=6cm]
{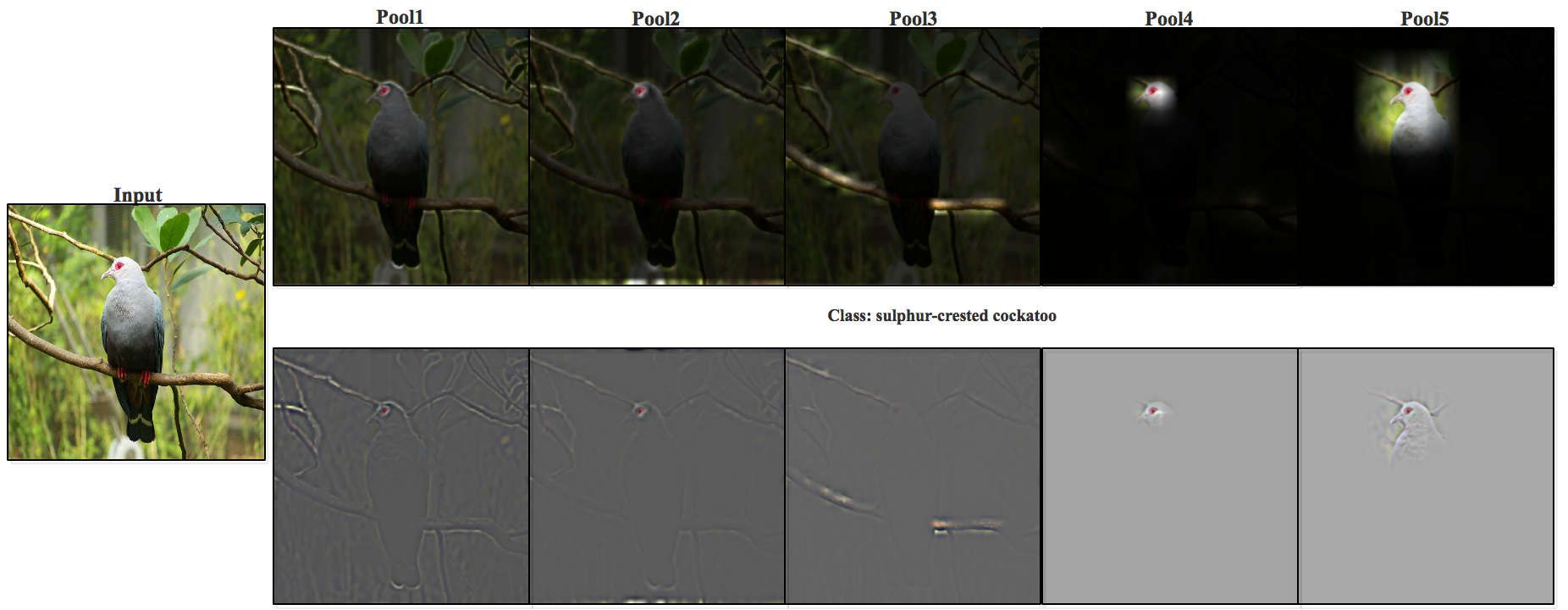}
\caption{Compressing information.}
\label{fig:compress}
\end{figure}

A lot of motivation for explanatory function arises from sensitive domains like medical diagnosis. In the case of this DCNN example, we need to understand the process of propagating information to the output layer. Our aim is to understand how attention changes as we propagate forward. The result shown in Figure \ref{fig:compress} explains the compression step, i.e. we can easily observe how the irrelevant information is filtered.

\textit{Understanding DCNN decisions in medical domain}

\begin{figure}[H]
\centering
\includegraphics[width=13.5cm,height=6cm]
{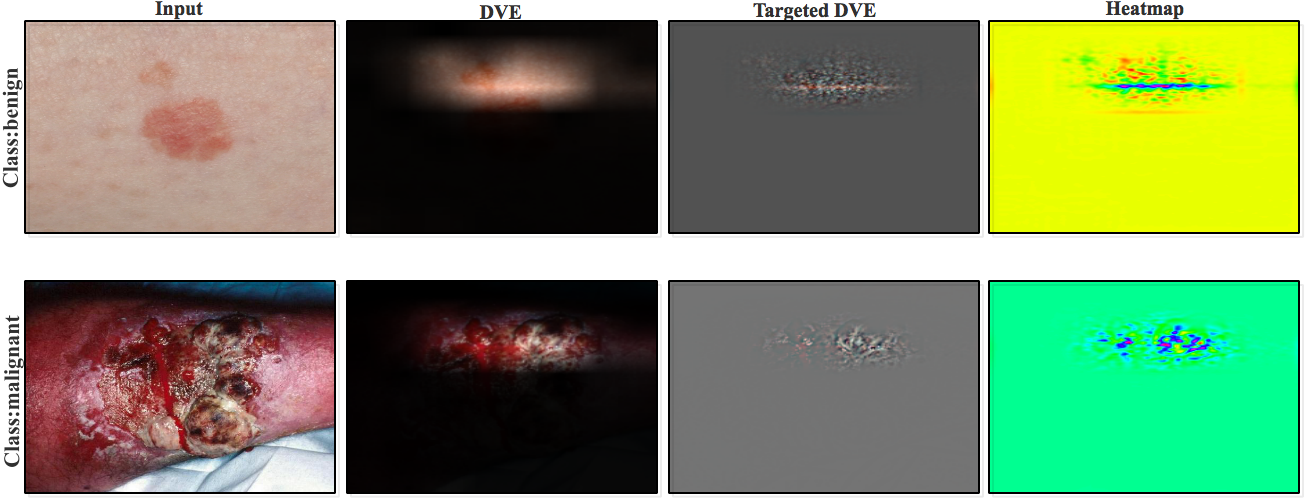}
\caption{Visualizing the support of the correct prediction.}
\label{fig:medical}
\end{figure}

Explaining DCNN prediction in medical domain is also important, because any decisions could have an impact on people's lives.  To show the effectiveness of our method, we used a pre-trained model [1] for skin lesion cancer classification, i.e. benign or malignant. The initial results of the visual explanation are depicted in Figure \ref{fig:medical}, we can see how the network is focusing on the most sensitive region in the image to make a decision.
\section{Conclusion}
We introduce a new framework for identifying explanations from DCNN decisions. Our approach captures the discriminative pixels by considering the activation of high and low spatial scales in Fourier space.   We experimented with a simple version of our approach on image classification. We also experimented with one of the potential applications of interoperability which is explaining predictions made for medical data.

\section{Future work}
The extension of this simple framework to other domains will help determine how the framework can be extended to more sophisticated domains, and more complex but interpretable explanations.  For example, within the more general framework of abductive explanation (e.g., [4]), explanations about the classification of language segments must include linguistic context (cf. discriminative pixels), or in speech, require audio context.  But the overall goal of providing context of a trained partial model and the identification of plausible components that give rise to specific classification output is the same.  In this way, we believe that the debugging of complex multi-dimensional learned neural network models will not just exhibit good performance, but can be debugged and interpreted to rationally improve performance.

\section{Acknowledgment} 

We thank colleagues from the Alberta Machine Intelligence Institute for their discussions and advice. This work was supported by NSERC and AMII.

\section*{References}
\medskip

\small
[1] Danish Shah, Jawad Shaikh, Afzal Sayed, Aditya Mishra, and Maaz Khan, https://github.com/DanishShah/DeepDiagnosis.

[2] Feichtenhofer C., Pinz A. and Wildes R. Spatiotemporal residual networks for video action recognition. In  \textit{Proc. NIPS} 2016 pages 3468-3476, 2016.

[3] Fong R., and Vedaldi A. Interpretable Explanations of Black Boxes by Meaningful Perturbation. arXiv preprint arXiv:1704.03296, 2017.

[4] G. Harman, Inference to the best explanation, $http://www.informationphilosopher.com/knowledge/best_explanation.html$

[5] LeCun Y., Bengio Y., and Hinton G. Deep learning. \textit{Nature} pages 436-444, 2015.

[6] Lin T.Y., Maire M., Belongie S., Hays J. Perona, P. Ramanan D., Dollár P. and Zitnick C.L. Microsoft coco: Common objects in context. In \textit{ECCV}, 2014.

[7] Mahendran A. and Vedaldi A.  Salient deconvolutional networks. In \textit{Proc. ECCV}, pages 120-135, 2016.

[8] Milan A., Rezatofighi S. H., Dick A. R., Reid, I. D. and Schindler K. Online Multi-Target Tracking Using Recurrent Neural Networks. In \textit{Proc. AAAI} AAAI pages 4225-4232, 2016.

[9] Rastegari M., Ordonez V., Redmon J., and Farhadi A.  Xnor-net: Imagenet classification using binary convolutional neural networks. \textit{Proc. ECCV} pages 525-542, 2016.

[10] Selvaraju R. R., Das A., Vedantam R., Cogswell M. Parikh, D. and Batra D. Grad-CAM: Why did you say that?. In \textit{Workshop on Interpretable Machine Learning in Complex Systems, NIPS}, 2016.

[11] Serban I. V., Klinger T., Tesauro G., Talamadupula K., Zhou B., Bengio Y. and  Courville A. C. Multiresolution Recurrent Neural Networks: An Application to Dialogue Response Generation. In \textit{Proc. AAAI} pages 3288-3294, 2017.

[12] Shelhamer E., Long J. and Darrell T. Fully convolutional networks for semantic segmentation. In \textit{IEEE transactions on pattern analysis and machine intelligence} pages 640-651, 2017.

[13]  Simonyan K., and Zisserman A. Very deep convolutional networks for large-scale image recognition. \textit{arXiv preprint arXiv:1409.1556}, 2014.

[14] Springenberg J. T., Dosovitskiy A., Brox T., and Riedmiller M. (2014). Striving for simplicity: The all convolutional net. \textit{arXiv preprint arXiv:1412.6806 }, 2014.

[15] Su B. and Lu, S.  Accurate recognition of words in scenes without character segmentation using recurrent neural network. In \textit{Pattern Recognition} pages 397-405, 2017.

[16] Szegedy C., Ioffe S., Vanhoucke V. and Alemi A. A. Inception-v4, Inception-ResNet and the Impact of Residual Connections on Learning. In \textit{Proc. AAAI} pages 4278-4284, 2017.

[17] Zeiler, M. D., \& Fergus, R. (2014, September). Visualizing and understanding convolutional networks. In European conference on computer vision (pp. 818-833). Springer, Cham.

[18] Zhou B., Khosla A., Lapedriza, A. Oliva, A. and Torralba A. (2016). Learning deep features for discriminative localization. In \textit{Proc. CVPR}, pages 2921-2929, 2016.

[19] Zintgraf L. M., Cohen T. S., Adel T., and Welling, M. Visualizing deep neural network decisions: Prediction difference analysis. In \textit{Proc. ICLR}, 2017.

[20] Montavon G., Lapuschkin S., Binder A., Samek W. and Müller, K. R. Explaining nonlinear classification decisions with deep taylor decomposition. In \textit{ Pattern Recognition} pages 211-222, 2017.

\end{document}